\documentclass[preprints,article,accept,moreauthors,pdftex]{Definitions/mdpi}

\firstpage{1} 
\pubvolume{1}
\issuenum{1}
\articlenumber{0}
\pubyear{2024}
\copyrightyear{2024}
\datereceived{ } 
\daterevised{ } % Comment out if no revised date
\dateaccepted{ } 
\datepublished{ } 
\hreflink{https://doi.org/} % If needed use \linebreak
\pdfoutput=1 % Uncommented for upload to arXiv.org
\Title{Quantum-Cognitive Neural Networks: Assessing Confidence and Uncertainty with Human Decision-Making Simulations}

\TitleCitation{Quantum-Cognitive Neural Networks: Assessing Confidence and Uncertainty with Human Decision-Making Simulations}

\definecolor{my_green}{rgb}{0.55, 0.71, 0.0}

\usepackage{comment}

\Author{Milan Maksimovic\orcidA{} and Ivan S.~Maksymov\orcidB{}$^{*}$}

\AuthorNames{Milan Maksimovic and Ivan S.~Maksymov}

\AuthorCitation{Maksimovic, M.; Maksymov, I.~S.}
\address{%
Artificial Intelligence and Cyber Futures Institute, Charles Sturt University, Bathurst, NSW 2795, Australia.}

\corres{Correspondence: mmaksimovic@csu.edu.au, imaksymov@csu.edu.au}

\abstract{Modern machine learning (ML) systems excel in recognising and classifying images with remarkable accuracy. However, like many computer software systems, they can fail by generating confusing or erroneous outputs or by deferring to human operators to interpret the results and make final decisions. In this paper, we employ the recently proposed quantum-tunnelling neural networks (QT-NNs), inspired by human brain processes, alongside quantum cognition theory, to classify image datasets while emulating human perception and judgment. Our findings suggest that the QT-NN model provides compelling evidence of its potential to replicate human-like decision-making and outperform traditional ML algorithms.}


\begin{document}

%%%%%%%%%%%%%%%%%%%%%%%%%%%%%%%%%%%%%%%%%%
%\setcounter{section}{-1} %% Remove this when starting to work on the template.
\section{Introduction}
\subsection{Motivation and Literature Review}
Uncertainty refers to the absence of complete knowledge about the present state or the ability to predict future outcomes accurately \cite{Bus12, Sni19, Gu20}. Nature, being a complex, nonlinear and often chaotic system \cite{Str15}, presents challenges when we attempt to anticipate its behaviour. Uncertainty also limits our understanding of human behaviour and decision-making patterns \cite{Bus12, Sni19}, thereby fuelling anxiety \cite{Gu20} and contributing to a lack of confidence in critical situations where accurate predictions are essential \cite{Luc20}. This makes the study of uncertainty especially pertinent to fields like machine learning (ML) and artificial intelligence (AI)~\cite{Hul21, Gaw23}.

\begin{figure}[t]
\centering
\includegraphics[width=0.99\columnwidth]{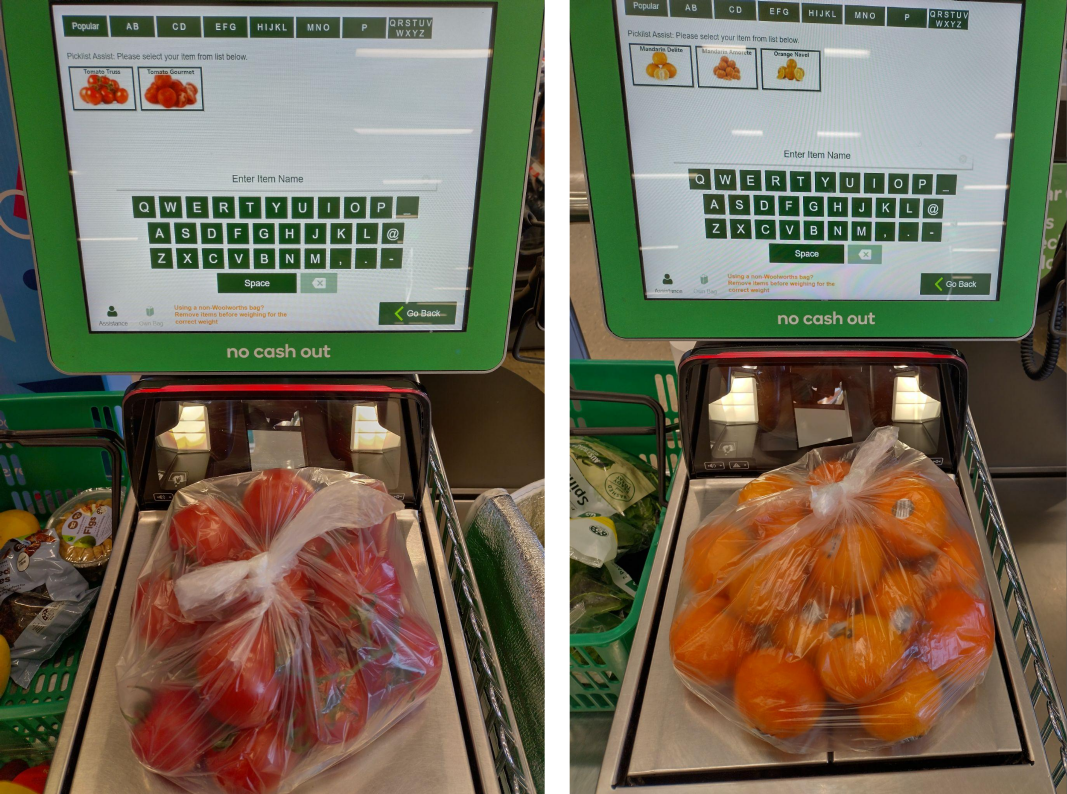}
\caption{Uncertainty in detecting fresh produce items at a supermarket self-checkout equipped with a machine vision system. Left:~The system analysed a transparent plastic bag containing truss tomatoes and identified two possible categories:~truss tomato and gourmet tomato, leaving the final selection to the customer. Right:~In another test with a bag of Amorette mandarins, the system suggested three potential options:~Delite mandarin, Amorette mandarin or Navel orange. Similar results were observed with other visually ambiguous items.}
\label{Fig0}
\end{figure}

Studies of uncertainty in AI and ML systems often draw on Shannon entropy (SE) \cite{Hul21, Guh23, Was23}---a fundamental concept in information theory that quantifies the uncertainty within a probability distribution \cite{Kar22}. SE specifically provides a mathematical framework for assessing unpredictability and information content associated with a random variable within an ML model, also establishing a link between natural intelligence and AI \cite{Sua20}. This figure-of-merit is important for understanding not only the internal behaviour of the model but also its confidence in making specific predictions \cite{Hul21, Was23}. For instance, SE can be applied to track changes in neural network weight distributions during training, where increased entropy may reflect heightened variability in weights as the model encounters complex or ambiguous data \cite{Hul21, Mak24_APL} (see Fig.~\ref{Fig0} for an example from real life). Entropy also enables distinguishing between confident and uncertain predictions in output distributions, where lower SE values often correspond to a higher confidence of the model in its classification outcomes \cite{Was23}. Additionally, SE has been useful in assessing uncertainty for probabilistic layers in neural networks and evaluating robustness in models exposed to noisy data \cite{Hul21}.

In traditional ML models based on artificial neural networks \cite{Hay98, Kim17}, the weights of network connections are derived from minimal information contained within the dataset \cite{Fra20}. However, such an optimisation often leads to poor model performance and limited generalisation capabilities \cite{Erh10, Jos22, Gaw23}. As a result, these models can not only lack accuracy but also frequently exhibit overconfidence in their predictions \cite{Wan21_1, Wei22}. This poses a significant risk in critical applications such as self-driving vehicles \cite{Mel22}, medical diagnosis \cite{Wan21_1} and financial modelling \cite{Bou23}, where failures could have severe consequences.

As a result, several approaches have been proposed to address this risk \cite{Muk23, Wei23}, including traditionally designed confidence-aware deep neural networks (DNNs) \cite{Ngu15, Guo17, Moo20, Wan23, Raf24}, Bayesian neural networks (BNNs) \cite{Jos22, Liu22, Gaw23, Was23} and various classes of quantum neural networks (QNNs) \cite{Wan17, Bee20, Yan21, Zha21, Cho24, Hie24, Pir24, Per24_2}. These models provide robust frameworks for developing uncertainty-aware neural networks, enhancing the reliability and safety of AI systems in high-stakes applications. 

The Bayesian approach in statistics contrasts with the frequentist perspective, particularly in how it handles uncertainty and hypothesis testing \cite{Sch21_1}. Its application in ML, especially in deep learning systems, offers several advantages over traditional methods \cite{Jos22}. These include improved calibration and uncertainty quantification, the ability to distinguish between epistemic and aleatoric uncertainty \cite{Hul21, Was23}, and increased integration of prior knowledge into models \cite{Jos22}. Architecturally, BNNs are typically categorised as stochastic models, where uncertainty is represented either through probability distributions over the activations or over the weights of the network \cite{Jos22}.

In QNNs, the concept of weights and activation functions takes on a distinctive interpretation compared to classical neural networks \cite{Mon21, Pir24, Per24_2}. In particular, their weights are typically represented as unitary transformations that evolve as quantum states and are governed by quantum bits (qubits) \cite{Pan23}. Unlike classical weights, which scale and adjust input values linearly or nonlinearly \cite{Kim17}, quantum weights manipulate complex probability amplitudes of qubits, leveraging superposition and entanglement to achieve richer representations of data \cite{Qiu19, Mon21, Bai24}.

The activation function, another key component in traditional neural networks \cite{Hay98}, presents unique challenges in the quantum realm due to the linear nature of quantum mechanics \cite{Nie02}. Indeed, nonlinear activation functions, such as rectified linear unit (ReLU) that is crucial for capturing complex patterns in classical systems \cite{Hay98}, are not directly implementable in quantum circuits \cite{Nie02}. However, a number of strategies have been devised to simulate nonlinear processes through measurement and interference processes \cite{Mar22, Per24_2, Par22}. These quantum-inspired activation mechanisms enable QNNs to handle complex datasets and perform tasks similarly to their classical counterparts, yet with extra functionality brought by quantum parallelism and entanglement \cite{Qiu19, Per24_2}.

The principles of neuromorphic computing can also be applied to build QNNs. Neuromorphic computers mimic the operational dynamics of a biological brain \cite{Tan19, Mar20, Mak23_review}. Various classical neuromorphic neural network models have been developed \cite{One22, Ye23}, which, rather than using conventional Boolean logic, process information through the nonlinear dynamic properties of physical systems \cite{Mak23_review, Mar20}.
\begin{figure}[t]
\centering
\includegraphics[width=0.7\columnwidth]{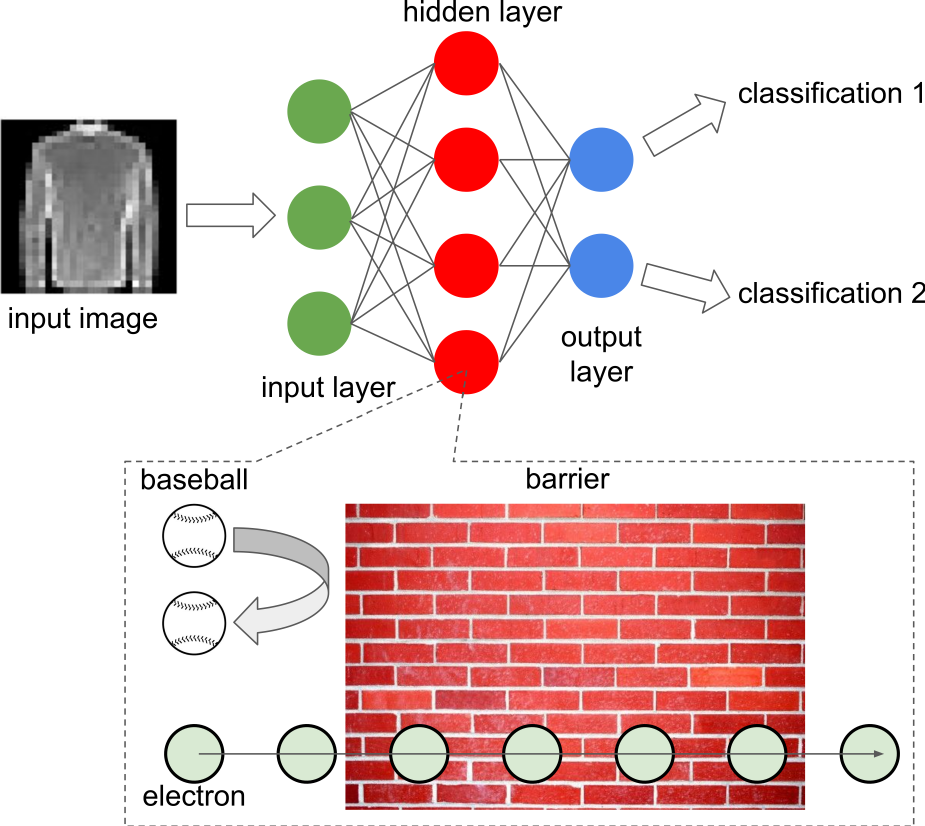}
\caption{Schematic representation of the QT-NN architecture. The inset illustrates the effect of quantum tunnelling that is employed as an activation function of the network.}
\label{Fig1}
\end{figure}

Recently, a novel neuromorphic QNN architecture was introduced \cite{Mak24_APL}, utilising the physical effect of quantum tunnelling (QT), which describes the transmission of particles through a high potential barrier \cite{McQ97, Gri04}. In classical mechanics, a baseball with energy $E<V_0$, where $V_0$ is the height of the barrier, cannot penetrate the barrier (see Figure~\ref{Fig1}, where the barrier is depicted as a brick wall). However, an electron, a quantum particle behaving as a matter wave, has a non-zero probability of penetrating the barrier and continuing its motion on the other side. Similarly, for $E>V_0$, the electron may still be reflected from the barrier with a non-zero probability.

Consequently, the neural network can be conceptualised as an electronic circuit, where connection weights correspond to the energies of electrons traversing the connections of the circuit. Importantly, these weights are updated according to the principles of quantum mechanics. In this framework, neurons do not require a specific accumulation of weight for activation since, instead, they function collectively in a probabilistic manner. The so-designed operating principle introduces additional degrees of freedom into the network, enhancing its flexibility and potential for processing of complex data~\cite{Mak24_APL}.

The quantum tunnelling neural network (QT-NN) model incorporates certain characteristics of BNNs and quantum computing-based QNNs. While the QT-NN does not utilise qubits, the application of the QT effect introduces additional degrees of freedom during network training, offering advantages akin to those provided by quantum parallelism and entanglement. At the same time, akin to BNNs \cite{Liu22, Ye23}, the QT-NN is a stochastic model that captures uncertainty through the probabilistic nature of its activation functions \cite{Mak24_APL} as well as via the injection of white noise and its application for the initialisation of connection weights \cite{Mak24_illusions}. 

However, the QT-NN model is uniquely different from any other competing approaches since it also incorporates the fundamental principles of quantum cognition theory (QCT) \cite{Atm04, Khr06, Bus12, Pot22}. Quantum cognition models offer a revolutionary perspective on understanding human decision-making and cognitive processes by integrating principles from quantum mechanics with cognitive psychology and decision-making theory \cite{Khr06, Bus12}. Unlike classical models, which often rely on deterministic frameworks \cite{Gal_book}, QCT postulates that human behaviour and perception of the world are inherently probabilistic \cite{Pot22}. For instance, QCT employs quantum superposition to explain how individuals can hold multiple, often contradictory beliefs and percepts simultaneously until a decision is made \cite{Bus12}. This approach not only enhances our understanding of cognitive phenomena, such as optical illusions, biases and uncertainty in judgment \cite{Bus12, Mak24_illusions, Mak24_APL, Mak24_information, Mak24_information1}, but also offers potential applications in the field of AI, where algorithms inspired by quantum cognition can improve decision-making processes in complex environments.

The QT-NN model is based on the mathematical solution of the Schrödinger equation \cite{Mak24_APL}. This equation is also pivotal for QCT due to its context-dependent solutions that provide insights into the complex patterns of human behaviour and perception \cite{Bus12, Mak24_information}. Importantly, the effect of QT has been naturally integrated into the fundamental framework of QCT \cite{Ben18, Mak24_illusions}, facilitating neural network models capable of capturing the intricate features of human behaviour. Furthermore, a strong connection has been established between QCT and models \cite{Geo18, Geo_book, Geo24} that attempt to explain human consciousness and brain function from the perspective of quantum information theory \cite{Chi19}. Consequently, the QT-NN serves as a powerful tool that leverages quantum physics, quantum information, psychology, neuroscience and decision-making, enabling the modelling of human choices with uncertainty in a unique way not available with other models.

\subsection{Objectives and Outline}
The objective of this paper is to investigate the potential of the QT-NN model in mimicking human-like perception and judgment. Drawing upon QCT, we aim to explore whether the QT-NN can replicate essential cognitive processes inherent in human decision-making, including ambiguity handling and contextual evaluation. Specifically, we examine whether the QT-NN can outperform conventional ML models in classifying image datasets, while also providing evidence of enhanced flexibility of the quantum approach and its ability to adapt to complex data patterns.

The remainder of this paper is structured as follows. We begin by introducing the theoretical foundations of the QT-NN model and its connection to QCT. Then, we present computational results demonstrating the advantages of the QT-NN, particularly its ability to provide human-like classification accuracy and flexibility. The following section discusses the implications of these findings, focusing on how the QT-NN model could enhance decision-making in critical real-world scenarios. Overall, by simulating human-like judgment, the QT-NN model aims to reduce the reliance on human operators and improve decision-making processes in situations involving uncertainty and complexity. 

\section{Methodology}
\subsection{Quantum-Tunnelling Neural Network}
In this section, we introduce the image recognition confidence model used throughout the remainder of the paper. The computational task of recognising and classifying images is a well-established challenge in the field of classical \cite{Kim17} and quantum ML \cite{Bai24}. We will use a generic neural network algorithm \cite{Has09} as a reference framework to highlight the essential algorithmic features and advantages of our proposed QT-NN model.

For the purposes of our study, we assume that all input images, for both training and testing, are greyscale with dimensions of 28$\times$28 pixels, aligning with the format of the MNIST and Fashion MNIST datasets discussed below. Hence, the QT-NN architecture includes an input layer with $L=28 \times 28$ nodes, a hidden layer containing $N = 800$ nodes and a final output layer with $M=10$ nodes for classification. Such a configuration has been shown to effectively predict MNIST images with an accuracy of less than 2\% \cite{Sim03}.

Weights between the nodes are generated using a pseudo-random number generator with a uniform distribution (for a relevant discussion see, e.g.,~\cite{Mak24_illusions, Mak24_APL}) and are updated during the training process using a back-propagation algorithm~\cite{Kim17}. The activation function $\phi_{QT}$ for the nodes in the hidden layers is defined using the algebraic expressions for the transmission coefficient $T$ of an electron penetrating a potential barrier. These expressions are well-established in quantum mechanics, with their final, ML-adopted algebraic forms detailed in the prior relevant publication \cite{Mak24_APL}.

The output nodes of QT-NN are governed by the Softmax function \cite{Kim17}
\begin{equation}
  \phi_{smax}(v_i) = \frac{\exp(v_i)}{\sum_{k=1}^{M} \exp(v_k)}\,,
  \label{eq:softmax}
\end{equation}
where $v_i$ is the weighted sum of input signals to the $i$th output node and $M$ is the number of the output nodes.

The network is trained and utilised as follows. First, we construct the output nodes that correspond to the correct classifications for the training datasets. The weights of the neural network are initialised randomly within the range from --1 to 1. After inputting the data $x_j$ and the corresponding training targets $d_i$, we compute the error $e_i$ between the output of the network $y_i$ and the target $d_i$ as $e_i = d_i - y_i$. 

Next, we propagate the error $\delta_i = e_i$ backward through the network, computing the respective parameters $\delta_i^{(n)}$ for each hidden node. This is done using the equations $e_i^{(n)} = W^{{(n)}^\top} \delta_i$ and $\delta_i^{(n)} = \phi_{QT}^\prime(v_i^{(n)}) e_i^{(n)}$, where $n$ denotes the layer number, $\phi_{QT}^\prime$ is the derivative of the activation function and $W^\top$ is the transpose of the weight matrix corresponding to that layer. This back-propagation process is repeated through each hidden layer until it reaches the first one.

Finally, we update the weights using the learning rule $w_{ij}^{(n)} \gets w_{ij}^{(n)} + \Delta w_{ij}^{(n)}$, where $w_{ij}^{(n)}$ are the weights between an output node $i$ and input node $j$ of layer $n$, and $\Delta w_{ij}^{(n)} = \alpha \delta_i^{(n)} x_j$. These steps are applied sequentially to all training data points, iteratively refining the weights to minimise the error across the dataset.

The above-outlined algorithm can be readily adapted to a classical framework by replacing the QT activation function with a standard ReLU activation function \cite{Kim17}. Relevant information can be found in the the prior publications \cite{Mak24_illusions, Mak24_APL}.

\subsection{Benchmarking Testbed}
Now we discuss the rationale behind our choice of test image dataset for this study. The MNIST \cite{Den12} and Fashion MNIST \cite{Xia17} datasets are widely used benchmarks in the field of ML, particularly for image classification tasks \cite{Moh20, Bai24}. The MNIST dataset consists of 70,000 greyscale images of handwritten digits (0--9) sized 28$\times$28 pixels. Its simplicity and accessibility have made it a standard testbed for developing and evaluating models \cite{Kim17}.

The Fashion MNIST dataset, introduced as a more challenging alternative \cite{Xia17}, contains the same number of greyscale images and structure but features 10~classes of fashion items, including shirts, shoes and bags, offering increased complexity over digits alone. These datasets serve as practical tools for training and testing ML models, providing insights into model accuracy, generalisation and robustness. Consequently, in the following analysis, we utilise the Fashion MNIST dataset.

Classifying images from the Fashion MNIST dataset presents a more challenging task for the chosen neural network architecture. However, we selected this dataset deliberately to increase the likelihood of neural network errors, mimicking situations in which humans make mistakes. Fashion MNIST has also been used in recent third-party work on QNNs \cite{Bai24}, enabling us to compare the outputs of our QT-NN model with the results produced by other quantum models described in the literature.

\subsection{Statistical Analysis}
Both outputs and weight distributions generated by the trained classical model and the QT-NN were formally compared using rigorous mathematical methods. Such techniques have often been used in the fields of classical and quantum ML \cite{Rud24}. Below, we provide the rationale behind our choice of the particular approaches. 

The Kullback–Leibler divergence (KLD) \cite{Csi75} is a well-established method in the field of ML for comparing the weight distributions of neural network models, as well as for conducting other types of statistical analysis \cite{Bur02, Wan23_1}. This approach quantifies how much information is lost when approximating the initial weight distribution with the trained weight distribution. A high KLD indicates that the training process has significantly altered the weight distribution, while a low KL divergence suggests the trained weights remain closer to the initial distribution. The results of KLD analysis are often visualised by comparing the probability density functions of the initial and trained weights, where the shaded area between them represents the divergence:~the larger the shaded area, the greater the difference between the distributions.

Alternative approaches include the Jensen–Shannon divergence (JSD) \cite{End03, Nie19}. A low JSD value indicates greater similarity between the two distributions, while higher values suggest more significant divergence. Unlike KLD, JSD is bounded between 0 and 1, providing a more intuitive measure of similarity. A value closer to 0 indicates that the distributions are highly similar, while values approaching 1 reflect substantial divergence. This provides a clearer indication of the differences, which justifies the use of JSD for the analysis in the remainder of this paper.

JSD is typically computed as~\cite{Nie19}
\begin{equation}
  JSD(p, q) = \frac{1}{2} \sum_{i=0}^{D} \left( p_i \log \frac{2 p_i}{p_i + q_i} + q_i \log \frac{2 q_i}{p_i + q_i} \right)\,,
  \label{eq:jsd}
\end{equation}
where $p$ and $q$ are the probability distributions, and $D$ is the dimension of the distributions. In this paper, it has been implemented using {\tt jensenshannon} procedure of the SciPy open-source Python library.
\begin{figure}[H]
\centering
\includegraphics[width=0.95\columnwidth]{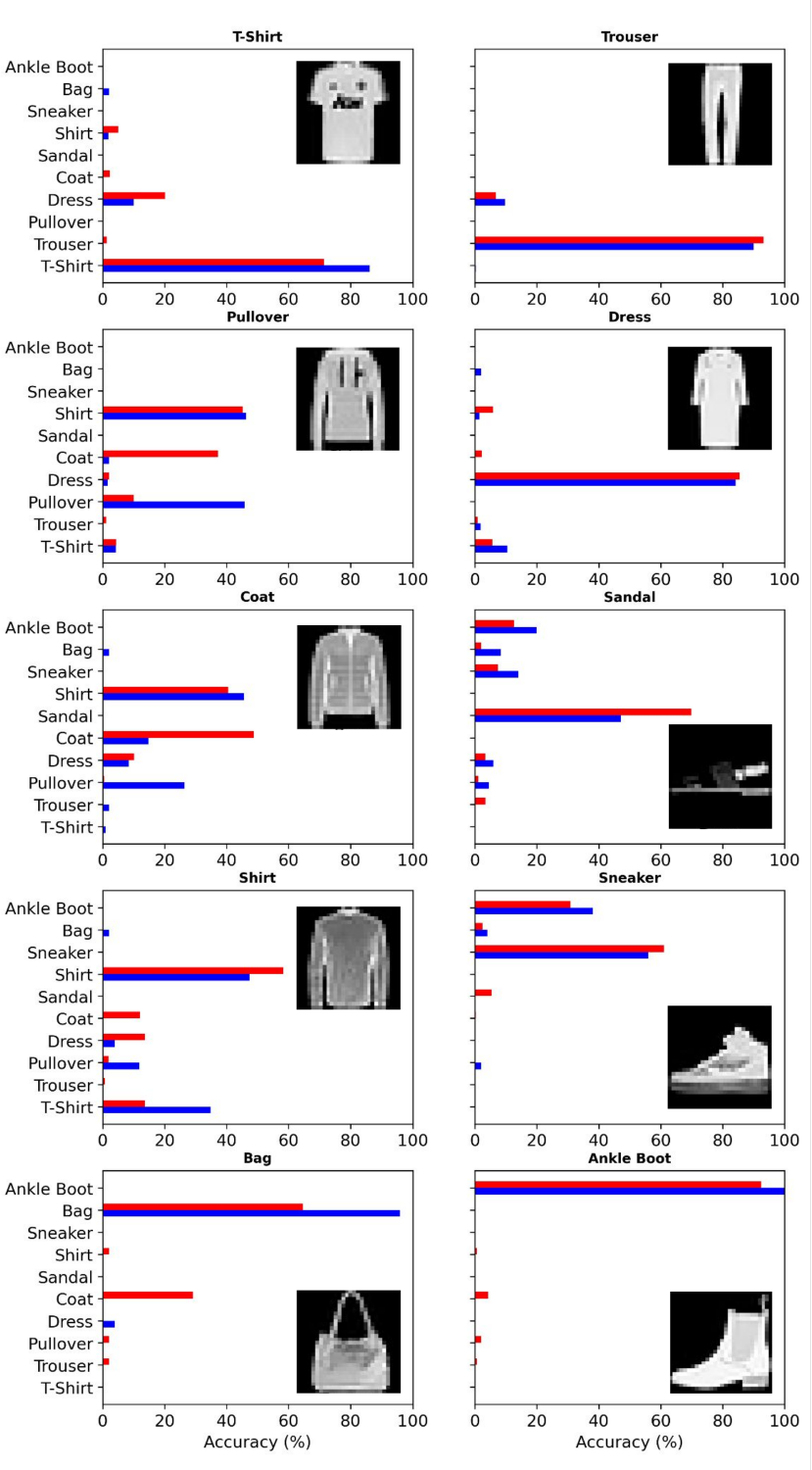}
\caption{Outputs generated by the QT-NN (red) and the classical neural network model (blue). The insets show the representative testing images for each classification category.}
\label{Fig2}
\end{figure}

\subsection{Model of Uncertainty}
Shannon entropy (SE) \cite{Kar22} is widely recognised as a standard model of uncertainty in studies of AI and ML systems \cite{Hul21, Guh23, Was23}. As a cornerstone of information theory, SE quantifies uncertainty within a probability distribution, providing insight into the variability of outcomes. SE is calculated as \cite{Was23}
\begin{equation}
H(p) = -\sum_{i=1}^{n} p(x_i) \log p(x_i)\,,
\end{equation}
where $p(x_i)$ represents the probability of occurrence of each possible outcome $x_i$ in a set of $n$ outcomes. Typically, a logarithm of base~2 is used, giving entropy values in bits. The negative sign ensures a non-negative entropy value since probabilities range between 0 and 1.

Essentially, SE reflects the uniformity of probability distributions:~when the probabilities are distributed more evenly, entropy is higher, indicating greater uncertainty about the outcome. Conversely, if one outcome dominates the probability distribution, the entropy value is lower, indicating less uncertainty. This principle underpins many applications of SE in AI and ML, where it helps assess model confidence and robustness by measuring the spread of predicted probabilities across possible outcomes~\cite{Hul21, Liu22, Was23}.

\section{Results}
\subsection{Comparison of the Outputs of QT-NN and the Classical Model}
In Figure~\ref{Fig2}, we compare the outputs generated by the QT-NN and a classical neural network model trained on the Fashion MNIST dataset.
To ensure a rigorous scientific comparison, both models were designed to have identical architectures, incorporating the same number of neurons, connections and initial random weight distributions, as well as identical training and testing procedures. Specifically, both models were trained on 32 batches of training image sets, with 100 training epochs for each batch---an optimal training strategy we identified---before classifying the same sequence of 50 test images. The respective outputs generated by the models were then averaged to produce the classification bar charts.

Figure~\ref{Fig2} provides an alternative to the commonly used confusion matrix approach \cite{Ste97}. We established that its composition offers more detailed information, which, in particular, has proven valuable in applying the statistical tools outlined earlier in the text. In particular, the title of each panel in Figure~\ref{Fig2} indicates the actual type of fashion item that each model was tasked to recognise and classify. The bar height represents the accuracy with which each model classifies the item. For example, when presented with 50 images of the `Trouser' category, the QT-NN and classical models correctly classify the items as `Trouser' with accuracies of 93.2\% and 90\%, respectively. Both models also suggest that these items could be 'Dress', with accuracies of 6.8\% and 10\%, respectively. As another example, while the classical model presented with the test images of the `Ankle Boot' category correctly identifies this item with accuracy of 100\%, the QT-NN model identifies it with 92.4\% accuracy, also suggesting that the items could be, with respective lower accuracies, `Coat', `Pullover', `Shirt' and `Trouser'.

The analysis of JSD and SE figures-of-merit provides insights into the prediction similarities and uncertainty levels between the QT-NN and the classical model. A lower JSD score, closer to 0, indicates high alignment between the probability distributions of the models, as seen in the `Trouser' category (JSD~=~0.049), where the two models produce nearly identical predictions. Conversely, a higher JSD score, like that of the `Pullover' category (JSD~=~0.403), signals substantial divergence, implying the models interpret this category differently. SE values supplement this by highlighting uncertainty levels in predictions, where a higher entropy indicates more ambiguity. Notably, the QT-NN model shows greater uncertainty for `T-Shirt' compared with the classical model (SE~=~0.8542 for the QT-NN compared with SE~=~0.5093 for the classical model). Additionally, categories like `Coat' and `Sandal' yield higher SE values for the classical model, suggesting that the predictions of the classical model are less confident. However, `Bag' has minimal entropy for the classical model (SE~=~0.1672), implying high certainty, whereas the QT-NN shows a moderate uncertainty (SE~=~0.8943).
\begin{figure}[t]
\centering
\includegraphics[width=0.99\columnwidth]{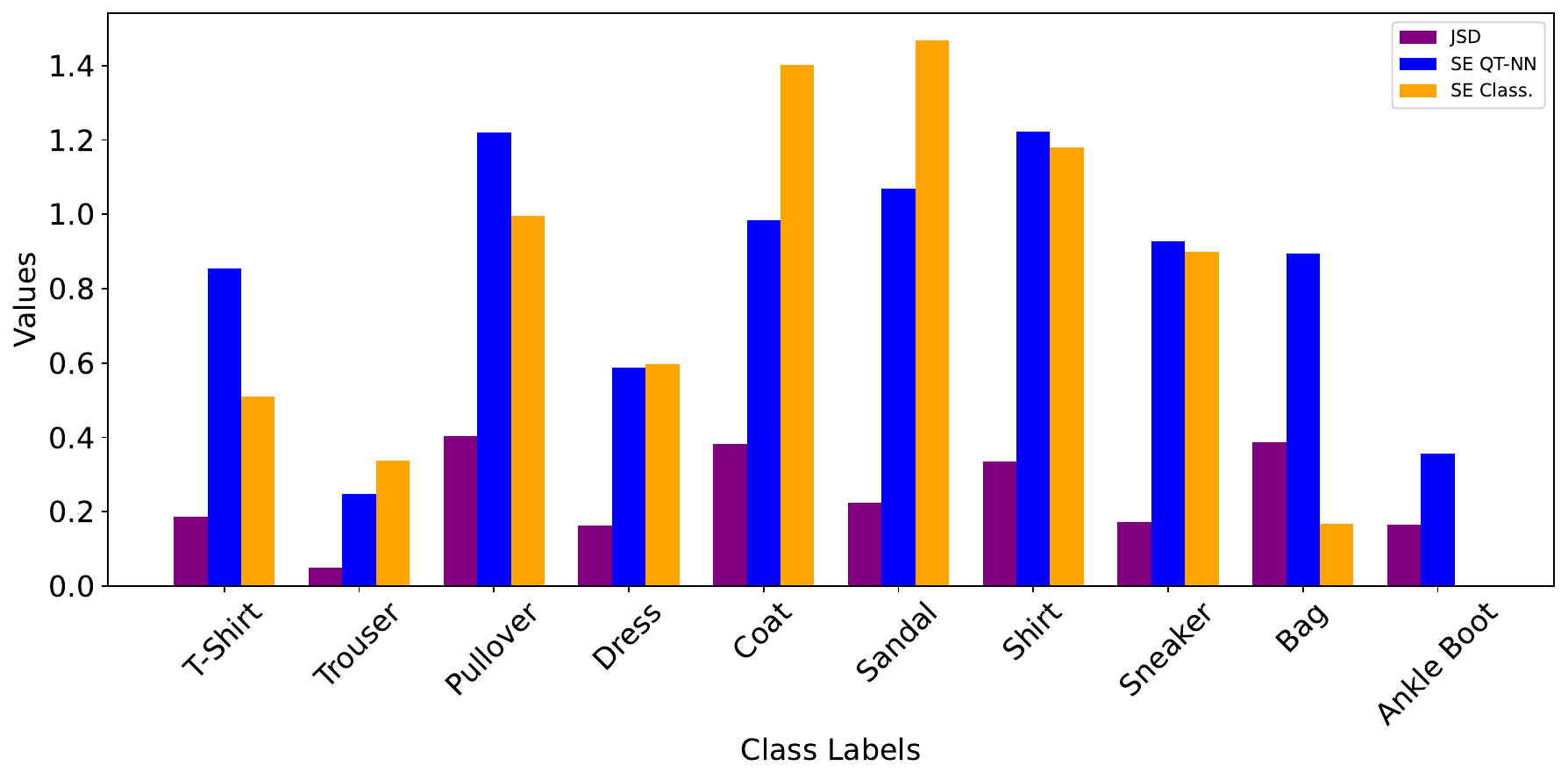}
\caption{JSD and SE figures-of-merit for the QT-NN and the classical model for each item category. Note that the classical SE is zero (to machine accuracy) for the `Ankle Boot' category.}
\label{Fig4}
\end{figure}

Thus, considering the overall ability of both QT-NN and classical models to accurately classify fashion objects---albeit with varying degrees of accuracy and similar patterns of failure---it becomes evident that the quantum model yields results that are more aligned with human cognition and perception of the world. Furthermore, given that the QT-NN achieves faster training times compared to the classical model (see the discussion below), we assert that it demonstrates superior overall performance. Additionally, the QT-NN compares favourably with more conventional superposition-enhanced quantum neural networks \cite{Bai24}, further underscoring its potential. 

Indeed, the outputs generated by both models are interpretable based on common human reasoning. Recalling that the images presented in the insets of Figure~\ref{Fig2} are representative examples of their respective categories and vary between batches, we observe certain pattern-related similarities among the categories of `Pullover', `Coat', and `Shirt'. Interestingly, images classified as `T-Shirt' share some common features with `Dress', while `Dress' exhibits similarities with `Trouser'. We also observe that images of footwear are distinctly recognised by both models as separate from the other categories. Conversely, `Bag' presents an intriguing test case since its contours may closely resemble those of a `Coat'.

In Section~\ref{Model_Sch}, we present an idealised, yet instructive, physical model that elucidates why the QT-NN is inherently more adept at managing the ambiguity of input data. This model demonstrates how the QT-NN produces results with higher confidence, mimicking certain essential aspects of human cognitive processes.

\subsection{Trained Weight Distribution Comparison}
In this section, we formally compare the weight distributions of the trained quantum and classical neural network models. We reveal that the QT-NN model can be trained up to 50~times faster than its classical counterpart. 

In traditional neural network training, entropy can serve as a figure-of-merit that provides insights into the level of order or disorder in the connection weights of the network. From the physical perspective, high entropy typically indicates a high degree of randomness and minimal structure within the network, which corresponds to a state of low energy or minimal work invested in training. Without significant training, the weights would remain in this disordered, high-entropy state. In contrast, effective training organises the weights into more structured distributions, where their values align with patterns that optimise input data processing \cite{Go04, Ngu17, Yos17, Ohz18, Fra20, Eil20}.

However, the QT-NN model differs fundamentally in its training approach. Indeed, rather than adjusting weights to fixed values, it aims to effectively use the entire range of possible weights represented by probability distributions. This approach reduces the need for extensive training since the QT-NN achieves optimal performance by incorporating the full spectrum of probable weights. Subsequently, the training process becomes more efficient and resource-effective, leveraging probabilistic weights to adapt dynamically rather than converging to fixed values. 

To demonstrate this, we treat the weight distribution matrices as two-dimensional random signals (essentially as white-noise images) and apply a fast Fourier transform to obtain their spectral information. (For simplicity, in the following, we denote the weights between the input layer and the hidden layer as $W_1$ and $W_2$ denotes the weights between the hidden layer and the output layer.) This approach enables us to analyse frequency components, revealing underlying patterns in the weight distributions that may not be apparent in the spatial domain. Then, using the {\tt histogram} function from the NumPy open-source Python library, which computes the occurrences of input data that fall within each bin, we plot the weight distribution. This computation is performed continuously throughout the training process, each time the neural network model is presented with new data. As a result, we obtain a plot showing the distribution of weights in the value range from --1 to 1 as a function of training iterations. The total number of training iterations is 32,000, corresponding to 32 batches of images trained over 100 epochs across 10 categories of fashion items.
\begin{figure}[t]
\centering
\includegraphics[width=1.05\columnwidth]{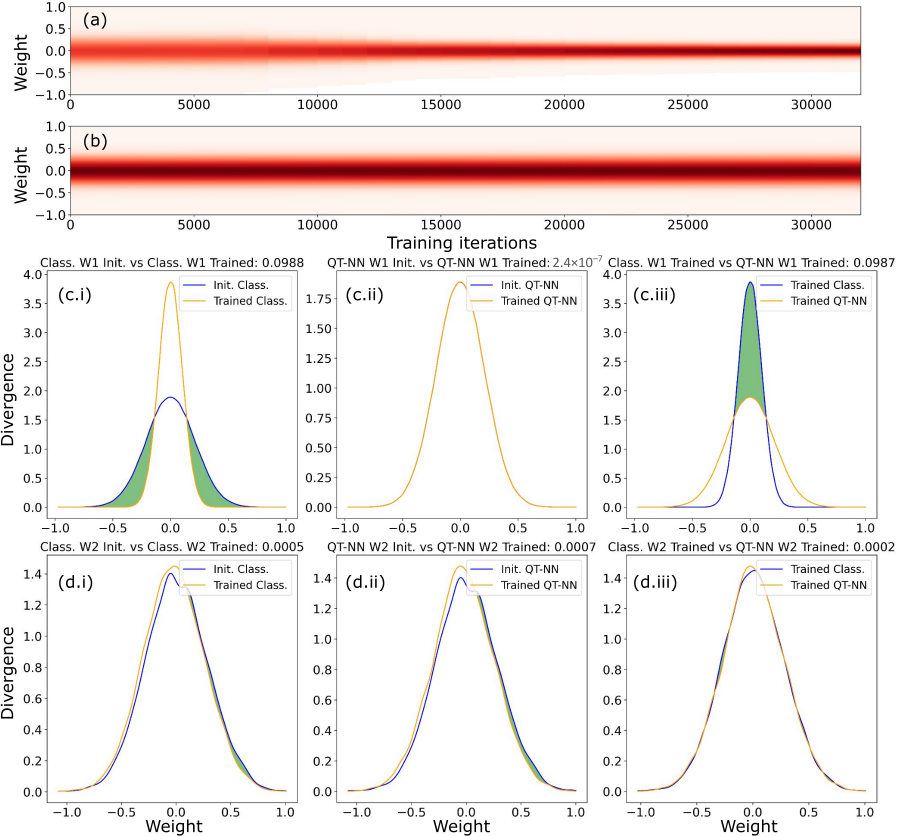}
\caption{{\bf(a,~b)}~Distributions of weights between the input layer and the hidden layer (denoted as $W_1$ in the main text), plotted as a function of training iterations for the QT-NN model and the classical model (labelled as 'Class.'), respectively. {\bf(c,~d)}~Results of the JSD cross-comparison of the initial (labelled as 'Init.') and trained weight distributions $W_1$ and $W_2$. The shaded areas in the JSD plots quantify the divergence, with the numerical value presented above each panel.}
\label{Fig3}
\end{figure}

In Figure~\ref{Fig3}a,b, the weight distributions exhibit Gaussian profiles centred around zero, which is consistent with the central limit theorem \cite{Bil95}. At the start of training, the weight distributions for both the QT-NN and classical models are identical. However, focusing on the weights $W_1$ in Figure~\ref{Fig3}a, training causes the distribution profile of the classical model to narrow, indicating a reduction in the diversity of weight values. This trend is further illustrated in Figure~\ref{Fig3}c.i, which shows the JSD analysis results for the initial and trained weights of the classical model.

Importantly, the narrowing of the Gaussian profile does not occur in the QT-NN model, as shown in both Figure~\ref{Fig3}b and Figure~\ref{Fig3}c.ii. Additionally, Figure~\ref{Fig3}c.iii provides a direct JSD comparison of $W_1$ between the trained classical and QT-NN models, further confirming this key difference in the training processes.

It should be emphasised that the observed value of JSD~=~2.4$\times10^{-7}$ in Figure~\ref{Fig3}c.ii is small but not zero, indicating that the training process of the QT-NN led to only a minor adjustment in the weights relative to the initial random weight distribution. This characteristic not only distinguishes the behaviour of the QT-NN from that of the classical model, but also suggests that a relatively small number of training epochs is required to achieve a satisfactory level of classification accuracy. Indeed, we found that the QT-NN can be trained up to 50 times faster than the classical model, underscoring the efficiency of quantum mechanics-based neural network models for solving classification problems. In particular, our results demonstrate that combining the QT effect with the additional degrees of freedom for data representation and processing facilitated by the probabilistic nature of quantum mechanics, along with a random initialisation of the weight distributions, enables more efficient training and exploitation of the model \cite{Bia17}. 

A similar analysis was conducted on the evolution of the weights $W_2$ for the connections between the hidden and output layers (see the corresponding panels (i)--(iii) of Figure~\ref{Fig3}d). Since both the QT-NN and the classical model employ the Softmax activation function for the output, the behaviour observed in Figure~\ref{Fig3}d is similar for both models and, therefore, is of less relevance to the main discussion in this paper.

\section{Discussion}
\subsection{Modelling of Human Judgement\label{Model_Sch}}
\begin{figure}[t]
\centering
\includegraphics[width=1.0\columnwidth]{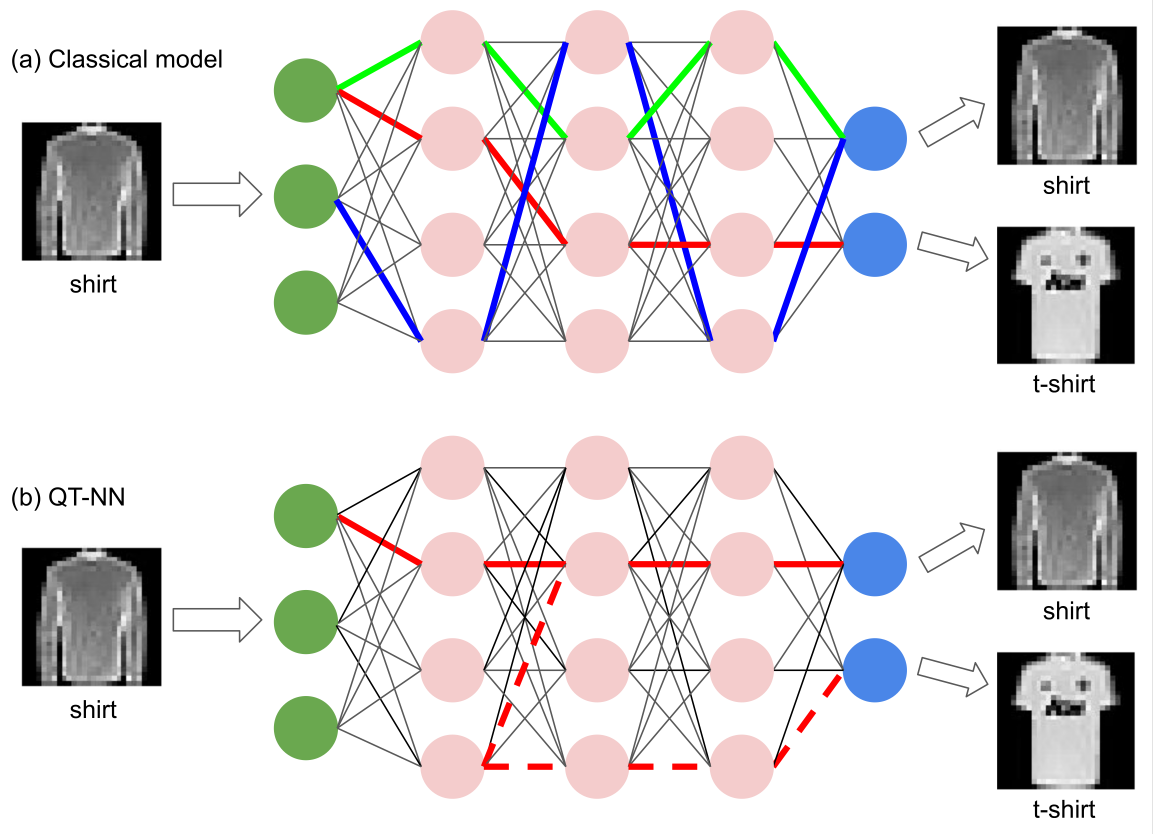}
\caption{Schematic illustration of the training process of {\bf (a)}~the classical model and {\bf (b)}~the QT-NN model, inspired by the discussion in Ref.~\cite{Yos17}. The coloured lines illustrate the possible pathways of neural connection formation. Note that the additional hidden layers of neurons are included purely for the sake of illustrating more advanced neural connections. Also note that the neural connections of the QT-NN model, depicted by the solid and dashed lines in panel~(b), are equally valid from the perspective of the algorithm and possess a probabilistic quantum nature.}
\label{Fig4_1}
\end{figure}

To establish a connection between the QT-NN model analysed in this paper and the group of interrelated theories collectively known as QCT, we extend the illustration in Figure~\ref{Fig1} by incorporating additional hidden layers into the artificial neural network (see Figure~\ref{Fig4_1}). This enhanced structure enables us to demonstrate that an idealised training process---similar to the natural learning mechanisms of a biological brain (see, e.g., Ref.~\cite{Bav12})---leads to the formation of robust neural connections that reflect how the neural network model organises and stores information \cite{Ngu17, Yos17}. 

Like a young child who needs to see numerous images of wild animals to distinguish them accurately, an artificial neural network must be exposed to many images of fashion items to achieve high classification accuracy. If the network is trained with only a limited number of shirt images, it can mistakenly develop an internal connection (illustrated by the red lines in Figure~\ref{Fig4_1}a) that causes it to classify the input as a t-shirt. However, through repeated training on the same category of inputs, the network forms a series of progressively more intricate connections (depicted by the green and blue lines in Figure~\ref{Fig4_1}a), enabling it to deliver correct classifications with high accuracy.

Continuing the analogy between the activation function of artificial neurons and a ball that must climb a brick wall to activate the neuron (Figure~\ref{Fig1}), we uncover a fundamental difference in the behaviour of the QT-NN model compared to the classical model. In the classical model, the ball can only overcome the barrier (the wall) when it accumulates sufficient energy. Translated to the connection weights within the network, this implies that the network must be exposed to the same category of input images multiple times. This repetition ensures that the sum of weights produces an equivalent energy value sufficient for the ball to surmount the barrier. 

This strategy appears to work effectively when the network is tasked with classifying unambiguous images. However, the natural or artificially introduced ambiguity of the inputs significantly complicates the task, prompting neural network developers to explore algorithms that allow for multiple classifications aligned with expert human judgment \cite{Wei23}.

Indeed, the approach based on the classical summation of weights has demonstrated certain limitations in modelling human perception of images that can have two or more possible interpretations \cite{Bus12, Ngu15}. A paradigmatic example of such images is the Necker cube, an optical illusion that alternates between two distinct three-dimensional perspectives depending on how the viewer interprets its edges and faces \cite{Kor05}. Alongside similar optical illusions, the phenomenon of the Necker cube illustrates the complexity of modelling perceptual ambiguity since human cognition switches between interpretations without requiring additional input or recalibration \cite{Ino94, Gae98, Kor05, Ben18, Ara20, Joo20, Mak24_illusions}.

Subsequently, it has been suggested that human judgment regarding the perceived state of the Necker cube can be effectively modelled using the principles of quantum mechanics \cite{Atm10, Bus12, Ben18, Mak24_illusions}. Specifically, it has been demonstrated that a general quantum oscillator model can plausibly simulate the switching of human perception between the two states of this optical illusion \cite{Bus12}. This approach provides a compelling framework for understanding the dynamic and probabilistic nature of human perceptual processes. Follow-up studies have demonstrated that incorporating a potential barrier into the quantum oscillator model allows for the capture of more intricate perceptual patterns, including extended periods of sustained perception of a single state of the cube \cite{Ben18, Mak24_illusions}. This enhancement provides deeper insights into the stability and transitions of human perception in the face of ambiguous stimuli \cite{Mak24_APL}.
\begin{figure}[t]
\centering
\includegraphics[width=1.0\columnwidth]{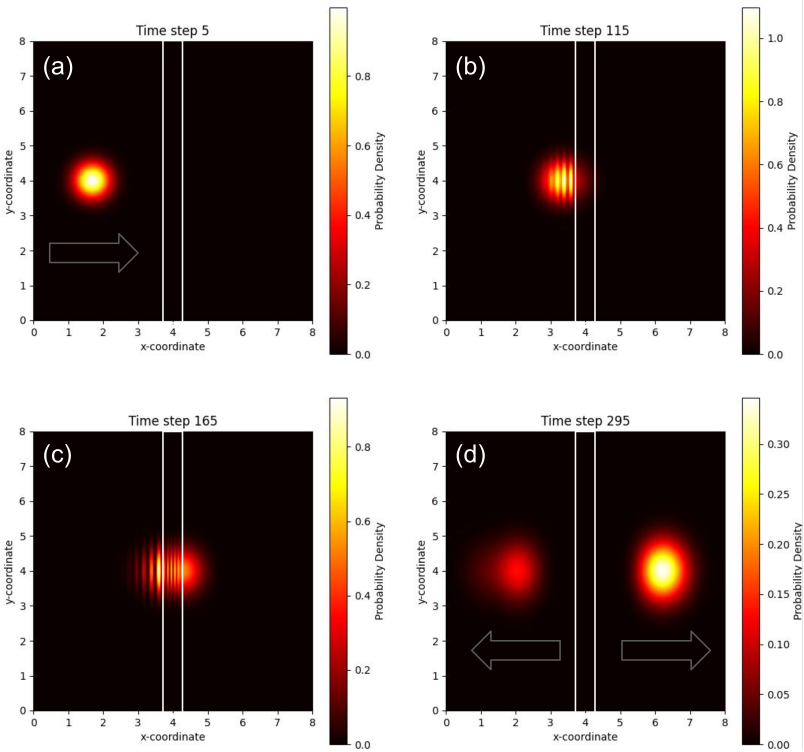}
\caption{{\bf(a--d)}~Instantaneous snapshots of an energy wave packet modelling the tunnelling of an electron through a potential barrier (depicted by a white rectangle). The false-colour scale of the images encodes the computed probability density values. Within the framework of the QT-NN model used in this paper, these values correspond to the connection weights of the neural network. The arrows indicate the direction of propagation of the wave packet. Note that in panel~(d), the wave packet splits into two parts, a physical phenomenon that would eventually lead to the formation of the connections illustrated by the solid and dashed lines in Figure~\ref{Fig4_1}b.}
\label{Fig5}
\end{figure}

Thus, as schematically shown in Figure~\ref{Fig4_1}b, the formation of the training link to the t-shirt (denoted by the dashed lines) no longer represents an algorithmic mistake for the QT-NN model. Indeed, visually, a shirt resembles a t-shirt, with the primary distinction being the absence of long sleeves in the latter. A similar perceptual ambiguity can be established between the other Fashion MNIST objects such as `Shirt', `Coat' and `Pullover'. This means that early in the training the QT-NN may form two possible pathways (denoted by the solid and dashed lines in Figure~\ref{Fig4_1}b) that are equally valid from the algorithmic point of view but have different probabilities to be triggered by the input image. The further training process primarily makes cosmetic refinements to these connections (in fact, the neural network can generate meaningful results as soon as its initial internal connections are established). This explains why the QT-NN model can be trained significantly faster than the classical model.

From the physical perspective, the process of tunnelling and probabilistic action of weight coefficients can be illustrated using a two-dimensional mathematical model, where the weights are represented as a Gaussian shaped energy packet that, in turn, represents a single electron that moves towards a potential barrier (Figure~\ref{Fig5}). The motion of the energy packet (the direction of which is indicated by the arrows in Figure~\ref{Fig5}) and its interaction with the barrier (indicated by the white rectangle in Figure~\ref{Fig5}) are governed by the Schr{\"o}dinger equation that is solved using a Crank-Nicolson method \cite{Kha22} (the same numerical experimental setup has been demonstrated to accurately model problems related to human cognition \cite{Mak24_book}). 

Plotting the probability density in two-dimensional space, Figure~\ref{Fig5} depicts the evolution of the energy packet through four snapshots taken at distinct instances of non-dimensionalised time. The peak amplitude of the initial packet (Figure~\ref{Fig5}a) corresponds to the sum of all weights associated with a single node in the neural network. The packet subsequently interacts with the barrier (Figures~\ref{Fig5}b and \ref{Fig5}c), producing both a reflected signal and a transmitted signal (Figure~\ref{Fig5}d).

The algebraic expressions underlying the operation of the QT-NN model \cite{Mak24_APL} encapsulate the same physical processes but do not offer the visual interpretability presented in Figures~\ref{Fig5}. However, in both physical interpretations, the thickness of the barrier serves as a hyperparameter that can be adjusted to control the model's confidence level in its assessment of fashion items. Moreover, the geometry of the barrier can be made more complex, for example incorporating a double-slit structure \cite{Mak24_book}. This complexity enhances the physics of interaction between the energy packet and the barrier, enriching the significance of this setup in the context of human cognition and decision-making modelling \cite{Bus12, Mak24_information, Mak24_information1}. Furthermore, it establishes a strong connection with competing QNN architectures being developed to enhance computational power and algorithmic advancements, particularly as tools for identifying patterns in data \cite{Bia17, Ohz18}.

\subsection{Practical Applications and Future Work}
Our future research work aims to explore and evaluate various classes of neural networks, including the QT-NN model and more traditional quantum and classical approaches. Each of these models, whether quantum-based or traditional, has unique strengths and weaknesses depending on the context in which they are applied. Therefore, it is important to acknowledge that no single model, including the QT-NN approach investigated in this paper, is universally superior. Moreover, significant experimental evidence gathered in recent years suggests that many urgent real-life problems can be effectively addressed mostly through hybrid approaches, combining the strengths of both classical and quantum models \cite{Lia21_1, Dom23, Gir23}.

Thus, subsequent exploration could investigate deeper how different neural network models make predictions and assess confidence levels. For instance, entropy measures on prediction outputs could highlight cases where quantum or classical algorithms overestimate their certainty. In such scenarios, hybrid classical-quantum models might intervene, potentially combined with human operator input, to refine and validate the outcomes, ensuring greater accuracy and reliability.

One particular avenue for future work is designing and developing hybrid quantum-Bayesian neural networks (QBNNs) \cite{Ngu22, Sak24}. These models can integrate the uncertainty quantification capability of Bayesian neural networks (BNNs) with the computational advantages of the QT-NN. A proposed hybrid QBNN architecture might employ a QT layer for initial feature extraction, followed by processing these quantum-derived features using a BNN layer, and concluding with either a quantum or classical layer for output generation. 

Moreover, these architectures could leverage quantum superposition and entanglement principles, enabling quantum layers to manipulate complex probability amplitudes. This approach could mimic human-like cognitive behaviour, such as evaluating multiple possibilities simultaneously before reaching a decision \cite{Bus12, Pot22, Mak24_information, Mak24_information1}. This hybridisation opens pathways to more nuanced and flexible AI models capable of addressing complex, real-world problems with improved reliability and decision-making capabilities.

In the fields of decision-making and QCT, superposition enables a model to consider multiple states at once, similar to how human cognition holds several options in mind before making a decision \cite{Khr06, Atm10, Bus12, Pot22, Mak24_information, Mak24_information1}. In a comparable way, the hybrid neural networks envisioned above should be able to mimic this behaviour by integrating new data and continuously refining decisions. Just as human subconscious processing helps weigh options and select the best course of action, hybrid neural networks would cross-reference different models to reduce overconfidence and incorrect AI predictions.

To support this, future research could adopt standard metrics such as entropy and divergence to measure uncertainty levels, positioning these hybrid models as a risk mitigation strategy for unreliable AI/ML predictions. Nevertheless, given the fast advancement of new technologies, novel approaches remain essential, and this paper has introduced several novel theoretical methods grounded in the principles of physics and engineering.

\section{Conclusions}
This study highlights the potential of the QT-NN model to mimic human-like perception and judgment, providing a promising alternative to traditional ML approaches. Using quantum cognition theory, our approach replicates cognitive processes observed in human decision-making, demonstrating that the QT-NN can classify image datasets with enhanced flexibility compared to conventional models. This result underscores the growing synergy between quantum technologies and AI, offering new avenues for AI systems to approach problems with more human-like reasoning and adaptability.

Our results not only open up new possibilities for AI development but also offer a pathway to refining decision-making in complex, real-world scenarios. While current ML systems generally excel in tasks like image recognition, they often require human intervention when faced with ambiguous or erroneous situations. The QT-NN model, as demonstrated, can bridge this gap by simulating human judgment, thereby reducing the reliance on human operators and potentially revolutionising fields where precision and nuanced decision-making are crucial, such as healthcare, autonomous systems and cognitive robotics.

%%%%%%%%%%%%%%%%%%%%%%%%%%%%%%%%%%%%%%%%%%
\vspace{6pt} 

%%%%%%%%%%%%%%%%%%%%%%%%%%%%%%%%%%%%%%%%%%
\authorcontributions{ISM developed the classical and quantum neural network models used in this work and obtained the primary results. MM conducted the statistical analysis and authored the respective sections of the article, including discussions on practical applications and future work. ISM edited the overall text with contributions from MM.}

\funding{This research received no external funding}

\institutionalreview{Not applicable.}

\informedconsent{Not applicable.}

\dataavailability{This article has no additional data.} 

%\acknowledgments{Not applicable}

\conflictsofinterest{The authors declare no conflicts of interest.} 

%%%%%%%%%%%%%%%%%%%%%%%%%%%%%%%%%%%%%%%%%%
%% Optional

%% Only for journal Encyclopedia
%\entrylink{The Link to this entry published on the encyclopedia platform.}

\abbreviations{Abbreviations}{
The following abbreviations are used in this manuscript:\\

\noindent 
\begin{tabular}{@{}ll}
AI & artificial intelligence\\
BNN & Bayesian neural network\\
DNN & deep neural network\\
JSD & Jensen–Shannon divergence\\
KLD & Kullback–Leibler divergence\\
ML & machine learning\\
MNIST & Modified National Institute of Standards and Technology database\\
QBNN & quantum-Bayesian neural network\\
QCT & quantum cognition theory\\
QNN & quantum neural network\\
QT & quantum tunnelling\\
QT-NN & quantum tunnelling neural network\\
ReLU & rectified linear unit\\
SE & Shannon entropy \\
\end{tabular}
}

%%%%%%%%%%%%%%%%%%%%%%%%%%%%%%%%%%%%%%%%%%
\begin{adjustwidth}{-\extralength}{0cm}

\reftitle{References}


\bibliography{refs}

%\PublishersNote{}
\end{adjustwidth}
\end{document}